\documentclass{article} 
\usepackage{nips12submit_e,times}

\usepackage{graphicx}
\usepackage{epsfig}
\usepackage{caption}
\usepackage{subcaption}
\usepackage{cite}
\usepackage{url}

\nipsfinaltrue

\begin{document}

\title{Clustering Learning for Robotic Vision}

\author{
Eugenio Culurciello\thanks{More information on Eugenio Culurciello's laboratory and research can be found here: http://engineering.purdue.edu/elab/. Real time robotic vision systems: http://www.neuflow.org/} \\
Purdue University\\
\texttt{euge@purdue.edu} \\
\And
Jordan Bates \\
Purdue University\\
\texttt{jtbates@purdue.edu}
\AND
Aysegul Dundar \\
Purdue University\\
\texttt{adundar@purdue.edu}
\AND
J.A. Perez-Carrasco \\
University of Seville \\
\texttt{jperez2@us.es}
\And
Clement Farabet \\
New York University \\
\texttt{cfarabet@nyu.edu}
}


\maketitle

\begin{abstract}
We present the clustering learning technique applied to multi-layer feedforward deep neural networks. We show that this unsupervised learning technique can compute network filters with only a few minutes and a much reduced set of parameters. The goal of this paper is to promote the technique for general-purpose robotic vision systems. We report its use in static image datasets and object tracking datasets. We show that networks trained with clustering learning can outperform large networks trained for many hours on complex datasets. 
\end{abstract}

\section{Introduction}

In the recent years the fusion of bio-inspired and neuromorphic vision models with machine learning has dominated the development of artificial vision systems for the categorization of multiple objects in static frames.
Bio-inspired deep networks are computer-vision and computational-neuroscience models of the mammalian visual system implemented in deep neural networks \cite{lecun_gradient-based_1998,hadsell_dimensionality_2006,gregor_structured_2011,riesenhuber_hierarchical_1999,serre_feedforward_2007,serre_neuromorphic_2010}. Most deep network architectures are composed of multiple layers (2, 3 typically), where each layer is composed of: linear two-dimensional filtering, non-linearity, pooling of data, output data normalization \cite{jarrett_what_2009,lecun_convolutional_2010,boureau_theoretical_2010}. 
Recent machine learning research has focused on the task of training such deep networks from the abundant digital data available in the form of image frames and videos. In particular, deep networks need to learn good feature representations for complex visual tasks such as object categorization and tracking of objects in space and time, identifying object presence and absence. These representations usually involve learning the linear filter weight values from labeled and unlabeled input data. Since labeled data is costly and often ridden with human errors \cite{karpathy_lessons_2011, torralba_unbiased_2011, hou_meta-theory_2012}, the recent focus is on learning these features purely from unlabeled input data \cite{olshausen_emergence_1996, hyvarinen_independent_2000, hinton_fast_2006, vincent_extracting_2008, coates_analysis_2011}. These recent methods typically learn multiple layers of deep networks by training several layers of features, one layer at a time, with varying complexity of learning models. 

Recent techniques based on unsupervised clustering algorithms are especially promising because they use simple learning methods that quickly converge \cite{coates_analysis_2011}. 
These algorithms are easy to setup and train and are especially suited for robotics research, because less complex knowledge of machine learning is needed, environment-specific data can be collected quickly with a few minutes of video, setup of custom size deep networks is quick and can be adapted to specific tasks. 
In addition, real-time operation with efficient networks can be obtained with only a few minutes of training and setup, leading to quick and direct experimentation in robotic experiments.

In this paper we present results obtained with unsupervised clustering algorithms on the training and operation of deep neural networks for real-time robotic vision systems. We provide simple techniques and open-source software that allows robotic researchers to use deep network in a short setup time and with little or no knowledge of machine learning necessary.
The main goal of the paper is not to present state-of-art results on a specific dataset. Rather we use standard published datasets to evaluate the performance of prototype robotic vision system for general-purpose use, where no dataset is available. It is thus not useful to train the network to perform only on one dataset, when the levels of performance would not carry over to another dataset or real-world images.
The goal is thus mainly to evaluate the use of unsupervised networks that can support at least ten frames-per-second operation on commercial hardware, such as recent laptop computers.

The paper presents the following key innovations: (1) use of clustering learning for quickly training robotic systems (Section \ref{sec-methods}), (2) the use of distance-based filtering, as opposed to the standard convolution (Section \ref{sec-net-arch}), (3) the experimental proof that clustering learning networks can outperform supervised networks on general purpose robotic tracking tasks (Section \ref{sec-results}).


\section{Methods}
\label{sec-methods}

In this paper we created and tested a model of unsupervised clustering algorithms (CL) that can quickly learn linear filters' weight values, and also is amenable to real-time operation with conventional mobile hardware.
We used the Torch7 software for all our experiments \cite{collobert_torch7_2011}, since this software can reduce training and learning of deep networks by 5-10 times compared to similar Matlab and Python tools.

\subsection{Input data}

Input image data was obtained from the CIFAR10 \cite{krizhevsky_learning_2009} and the Street View House Numbers (SVHN) \cite{netzer_reading_2011} datasets. The SVHN dataset has a training size of 73,257 32x32 images and test size of 26,032 32x32 images.
The CIFAR10 dataset has a training size of 20,000 32x32 images and a test size of 2,000 32x32 images. Both datasets offer a 10 categories classification task on 32x32 size images. The train dataset was in all cases learned to 100\% accuracy, and training was then stopped. Input data was contrast normalized separately on each RGB channel with a 9x9 gaussian filter using the Torch7 "nn.Spatial Contrastive Normalization" function.

For the real-time networks in Section \ref{section-realtime}, we used patches from a contrast-normalized version of a few images from the Berkeley image dataset \cite{martin_database_2001}.  As testing data for a general-purposed robotic vision system, we decided to test our networks in a tracking task, where the network has to be able to track an object of interest based on a  single presentation. For this purpose we use the challenging benchmark TLD dataset \cite{kalal_tracking_2012}. From this dataset we selected multiple videos with different properties of occlusions, camera movement, pose, scale and illumination changes.

Even if other groups showed slight improvements using the YUV color space \cite{jarrett_what_2009}, we were not able to reproduce the benefits of YUV, so we kept the images in their original RGB space. 

Also we did not use whitening of data (such as ZCA whitening) even if other groups have shown clear advantages of using it.  We did not use whitening because of two main reasons: first, it is not applicable for general-purpose vision system where an a-priori dataset cannot be obtained. Second, whitening computation is very expensive (similar to the first layer of a convolutional neural network) and we instead replaced it with local contrast normalization, which is a bio-inspired technique \cite{wandell_foundations_1995} to whiten the input data removing its mean bias and adapting the input dynamic range.

\subsection{Network architecture}
\label{sec-net-arch}

The deep neural network architecture is composed of 4 layers, not counting pooling and normalization operations. Two layers of linear two-dimensional filtering and two layers of output classifier in the form of a fully connected 2-layer neural network. The first two layers were composed of a two-dimensional convolutional linear filtering stage, a L2 norm pooling stage, and a subtractive normalization layer for removing the mean of all outputs. 
The filters of the first two layers are generated with unsupervised clustering algorithms, as explained below. Training of the last two fully-connected neural network layers was performed with approximately 50-100 epochs on the SVHN dataset on a quad-core Intel i7 laptop, or about 8 hours. Test data maximum precision usually only needed approximately 15 epochs.

\begin{figure}
\centering
\includegraphics[width=5in]{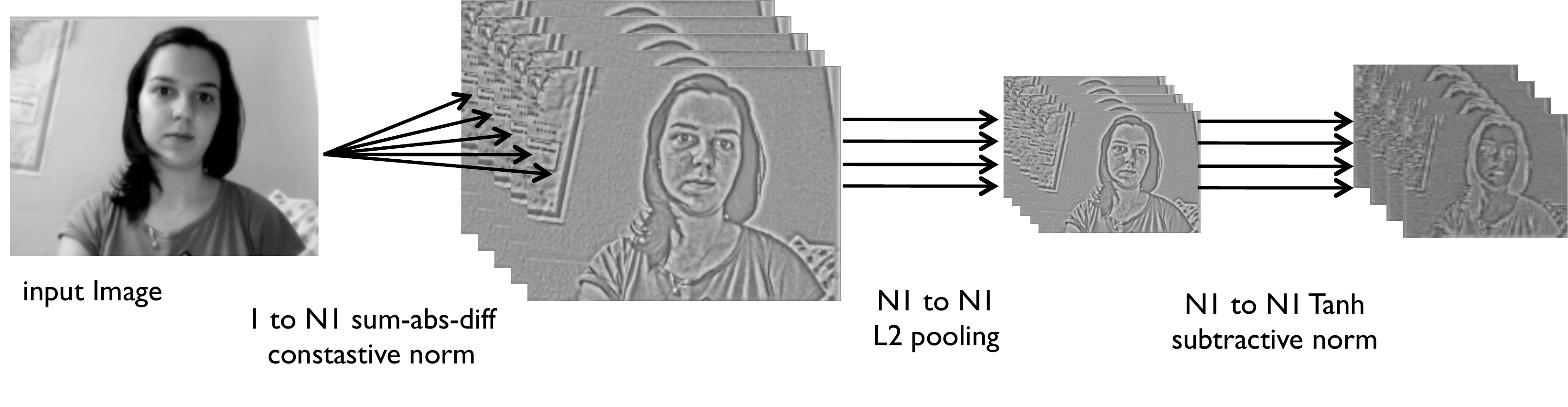}
\caption{Architecture of one layer of the clustering learning (CL) network. Filters were applied with a sum-abs-diff operations, followed by contrastive normalization, L-2 pooling of features, nonlinearity tanh, and subtractive normalization.}
\label{fig-CL-layer}
\end{figure}

The layers in the clustering learning network used the following sequence of operations. Using the naming convention in \cite{lecun_convolutional_2010}, for each layer $l$  $x_i$ is an input feature map, $y_i$ is an output feature map. The input of each layer is a 3D array with $n_l$ 2D feature maps of size $n_{l1} \cdot n_{l2}$. Each component (pixel, neuron) is denoted $x_{ijk}$. The output is also a 3D array, $y_i$ composed of $m_l$ feature maps of size $m_{l1} \cdot m_{l2}$.

\begin{enumerate}
\item SpatialSAD module: performing sum-abs-diff operation on images convolutionally with the learned CL filters: $yc_i=\sum_i |{k_{ij} - x_i}|$, where $|x|$ is the absolute value of $x$.
\item Spatial Contrastive Normalization: to zero the data mean and standard deviation
\item L2 pooling over 2x2 regions: $yp_i = \sum_{n \times n}(yc_{ij} \cdot gn_{ij})$, computing the weighted average of the input in an $n \cdot n$ (here $n=2$) region with a gaussian kernel $gn$.
\item Tanh nonlinearity: $ynl_i = tanh(yp_i )$
\item Spatial Subtractive or Contrastive Normalization: to zero the data mean, reset std to unity. The subtractive normalization operation for a given site $ynl_{ijk}$ computes: $v_{ijk} = ynl_{ijk} - \sum_{ipq} w_{pq} \cdot ynl_{i,j+p,k+q}$, where $w_{pq}$ is a normalized truncated Gaussian weighting window (typically of size 9x9). The divisive normalization computes $y_{ijk} = v_{ijk}/max(mean(\sigma_{jk}),\sigma_{jk})$ where $\sigma_{jk} = (\sum_{ipq} w_{pq} \cdot v^2_{i,j+p,k+q})^{1/2}$. The local contrast normalization layer is inspired by visual neuroscience models.
\end{enumerate}

In order to show the effectiveness of the learning techniques, we compared them to a standard 1-layer and a 2-layers convolutional neural network (CNN) \cite{sermanet_convolutional_2012,boureau_theoretical_2010,lecun_convolutional_2010}, followed by the same 2-layers classifier. The layers in the convolutional neural network used the following sequence of operations:
\begin{enumerate}
\item SpatialConvolution module: performing convolutions on images with the learned CL filters: $yc_i=b_j+\sum_i{k_{ij}\ast x_i}$, where $\ast$ is the 2D discrete convolution operator and $b_j$ is a trainable bias parameter.
\item L2 pooling over 2x2 regions: as described above.
\item Tanh nonlinearity: as described above.
\item Spatial Subtractive Normalization: as described above.
\end{enumerate}

All networks used 16 filters on the first layer, 128 filters on the second layer. The final classifier was fixed to 128 hidden units and 10 output classes for CIFAR and SVHN. Clustering learning networks used a fully connected (all channels connected to all channels) input to 1st, and 1st to 2nd layer, while convolutional neural networks used 1-out-of-3 and 4-out-of-16 random connection table between the input and 1st layer, and 1st and 2nd layer respectively.

Notice that CL networks used sum-abs-diff metrics to correlate filters responses to inputs. This is different to the standard approach of deep networks \cite{lecun_gradient-based_1998, hinton_fast_2006, krizhevsky_learning_2009, jarrett_what_2009} where convolution operations are used. Our choice of sum-abs-diff operations was dictated by improved performance of the CL filters with respect to convolutions.
In a standard modern computer the difference between convolutions (multiplications by weights) and distance metrics (differences, subtractions) is not visible, as multipliers are optimized to perform as fast as the simpler difference operations. On the other hand, when programmable hardware as FPGA is used, the use of differences instead of multipliers can reduce the silicon area utilization, and power up to 15 times on a 16 bit operation, and more for larger number of bits. A full comparison of the hardware advantages of using distance operators instead of convolutions will be the subject of future publications.

\subsection{Learning}
We use k-means clustering algorithm to learn a set of 16 filters in the first layer, and 128 filters in the second layer. The techniques and scripts are general and can quickly modified to learn any number of filters. The filter sizes on both layers was set to 5 x 5 pixels for the SVHN datasets. In CIFAR 1st layer we used 5 x 5 filters, and on the 2nd layer, we used 3 x 3 filters, as features were more spatially constrained in this dataset. Clustering used the same size patches of the normalized images, and we used 1 M patches from each dataset to train the first layer. The second layer training was performed by passing the entire dataset through the first layer of the deep neural network. The output dataset was then used again with the same script to train another set of linear filters, by using 1 M patches of the processed dataset.

\begin{figure}
        \centering
        \begin{subfigure}[b]{0.3\textwidth}
                \centering
                \includegraphics[width=1.2in]{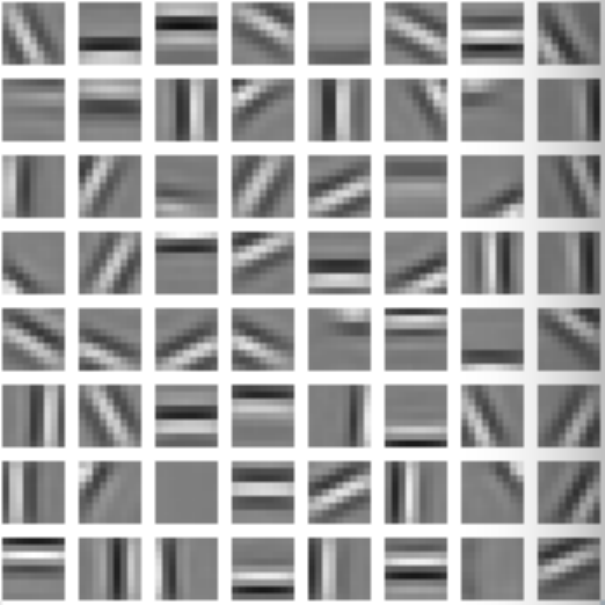}
                \caption{1st layer filters}
        \end{subfigure}%
        ~ 
        \begin{subfigure}[b]{0.7\textwidth}
                \centering
                \includegraphics[width=2.4in]{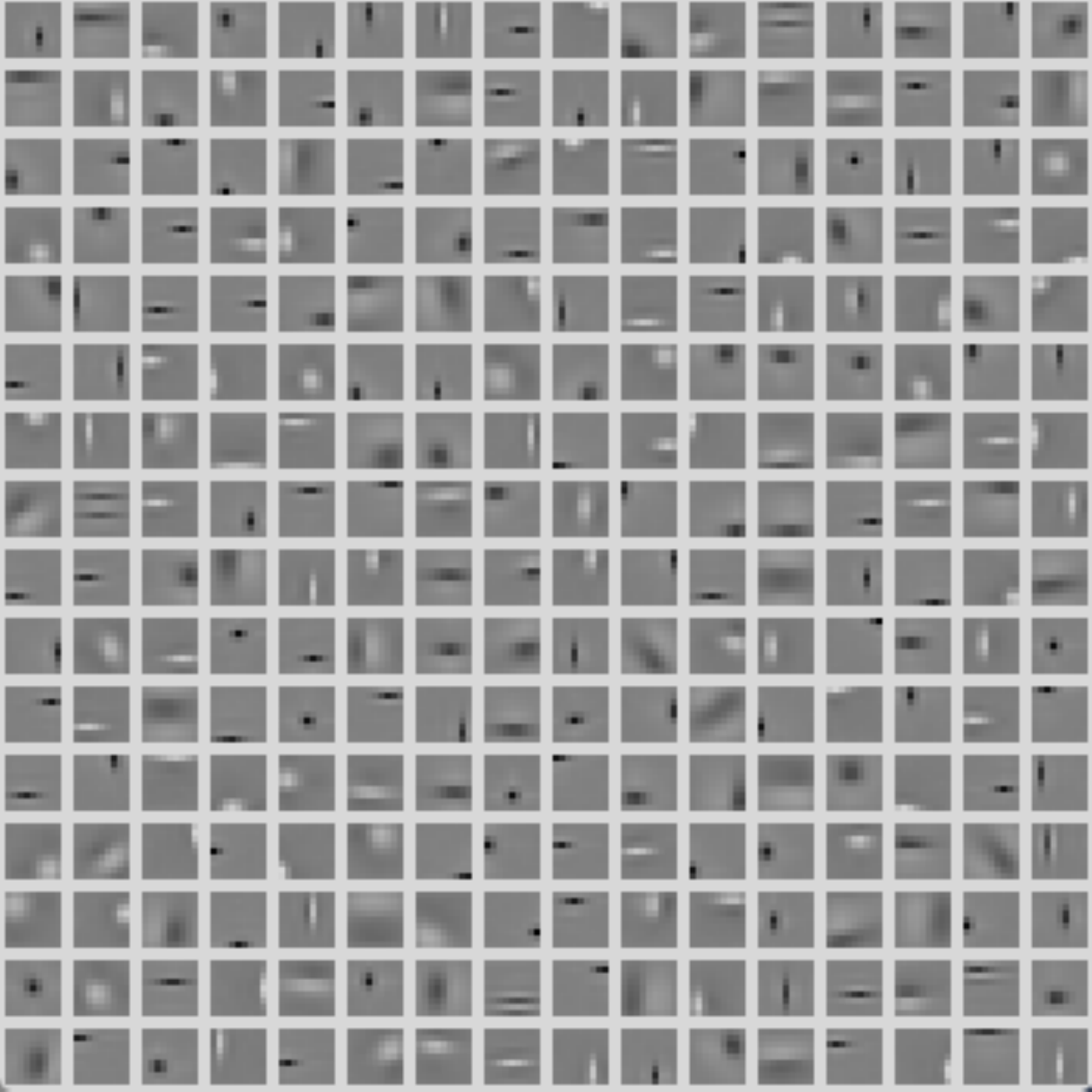}
                \caption{2nd layer filters}
        \end{subfigure}
        \caption{Filters obtained with clustering learning on the 1st and 2nd layer. The filters obtained on the 1st layer are quite similar to elongated Gabor patches, and what can be obtained with more complex and numerically involved unsupervised techniques. Filter training with CL was obtained in ~10 min time on a modern laptop.}\label{fig-filters}
\end{figure}

Clustering learning filters learned on the 2nd layer used as many planes as the output of the 1st layer (16 here). For example we learned 128 filters each with 16 dimensions of 5 x 5 pixels. This was done to cluster different sets of features for each output of the 1st layer, and increased performance by an average of 5\% or more.

Examples of the filters learned with CL techniques on a 1st and 2nd layer are given in figure \ref{fig-filters}. Both layer filters were obtained with training for ~10 minutes on a modern quad-core Intel i7 laptop. A supervised network of this size would require several tens of thousands of image examples and several hours of training time.

Convolutional neural networks were trained with supervised methods using stochastic gradient descent and other techniques as mentioned in \cite{lecun_convolutional_2010}.

\section{Real-time network}
\label{section-realtime}

The goal of this paper was to provide a simple and fast method to train unsupervised networks for general-purposed robotic vision system.  For real-time experiments we used the TLD dataset \cite{kalal_tracking_2012}. We then compared the tracking performance of a real-time deep network both trained supervised \cite{farabet_scene_2012}, and with CL techniques described here.

The CL network for this task had the same two-layer network architecture used in \cite{farabet_scene_2012}. This insured real-time operation of ~6 frames/s on a quad-code Intel i7 laptop computer. We focused on this network and restricted ourselves to real-time operation because the goal of this project is the use of deep networks in mobile computers. The network operates on 46 x 46 input images, uses 16 filters with 7 x 7 receptive field on the first layer and 128 filters with 7 x 7 receptive field on the second layer. The 1st to 2nd layer fan-in was 8, and the 2nd to 3rd layer fan-in was 64. Both layers were connected with random tables. The network produces a 128 feature vector as output. 

We trained the same size and number of filters through clustering algorithm for use in this network. We used patches from a contrast-normalized version of a few images from the Berkeley image dataset \cite{martin_database_2001}. Any set of natural-scene images can be used to train the general-purpose network presented in this section, and we on purpose chose not to sample patches from the target TLD dataset, in order to demonstrate the learning invariance properties of our technique.




\section{Results}
\label{sec-results}

\subsection{Static datasets}

\begin{figure}[htbp]
\begin{center}
\includegraphics[width=0.7\textwidth]{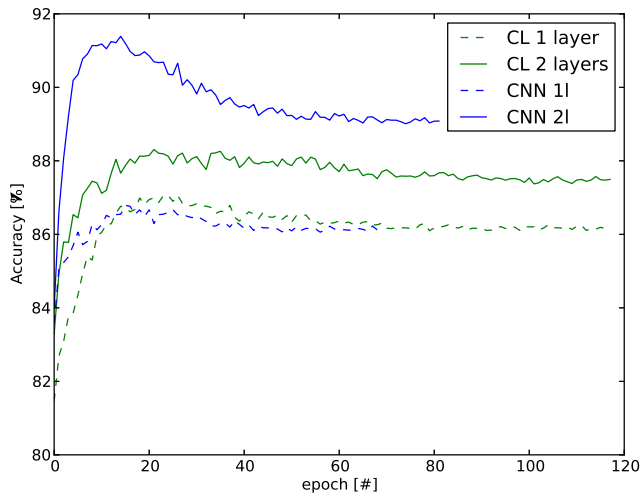}
\caption{Test set accuracy comparison for a convolutional neural network (CNN) and a clustering learning (CL) network with 1 and 2 layers on the SVHN dataset}
\label{data_svhn}
\end{center}
\end{figure}

We report the results in the SVHN dataset in figure \ref{data_svhn}. Here we compared results of accuracy in the test set for 4 cases: clustering learning with 1 layer (CL 1 layer), clustering learning with 2 layers (CL 2 layers), a 1-layer and a 2-layers convolutional neural network (CNN 1l, 2l). 

The data shows that clustering learning and CNN with 1 layer provide remarkably the same levels of accuracy on the dataset. This shows that at least on the 1st layer, the features learned with clustering learning are very effective.
Adding a second layer brings the convnet to 91\% levels of accuracy, which are standard without using any sophisticated tricks and with a small network with only 16 filters on the first layer.
On the other had, the clustering learning network with 2 layers showed more than 1\% increase in accuracy, from 88\% to 89\%. This increase is not as large as one would want and expect from adding a second layer, but is consistent with unsupervised learning results \cite{coates_analysis_2011,coates_selecting_2011}.

It is interesting to note that with the clustering learning 2 layers network accuracy was above 88\%, and plateaued with the train set accuracy plateauing also at 92\%. This shows that clustering learning filters also do not over fit, and present non-perfect, but almost identical results on both train and test sets. We note that CNN network trained supervised on SVHN report state-of-the-art performances of 96\% and above.

The results above were all obtained with feed-forward hierarchical networks. We also tried to use multiple layers of clustering learning unsupervised networks in parallel, as recommended by other publications \cite{coates_analysis_2011,lecun_convolutional_2010}, but we did not obtain any benefits from that strategy, on the contrary parallel networks always reported losses of 3-5\% accuracy with respect to a single layer. This is different from what reported in \cite{coates_analysis_2011,coates_selecting_2011}.

\begin{figure}[htbp]
\begin{center}
\includegraphics[width=0.7\textwidth]{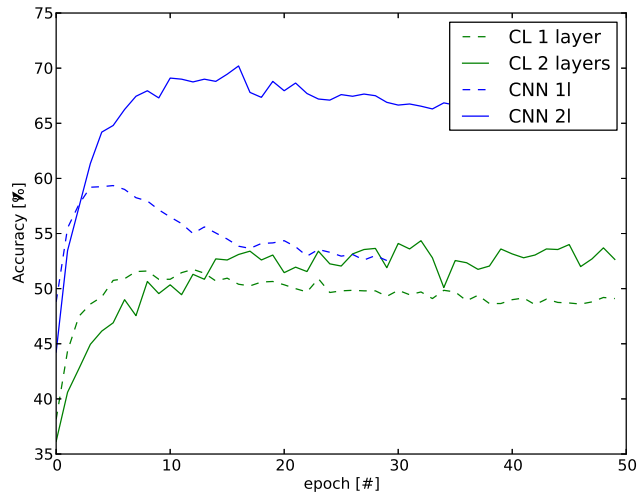}
\caption{Test set accuracy comparison for a convolutional neural network (CNN) and a clustering learning (CL) network with 1 and 2 layers on the CIFAR10 dataset}
\label{data_cifar}
\end{center}
\end{figure}

We also report here the results in the CIFAR10 dataset in figure \ref{data_cifar}. As in the SVHN case, we compared results of accuracy in the test set for 4 cases: clustering learning with 1 layer (CL 1 layer), clustering learning with 2 layers (CL 2 layers), a 1-layer and a 2-layers convolutional neural network (convnet 1l, 2l).

The results in the CIFAR10 dataset show a gain of more than 10\% from using a single layer convnet to a 2 layer convnet. 
Clustering learning showed the same behavior as in the SVHN dataset: adding a second layer achieved 3-5\% better accuracy on the test set. This dataset uses very small images and is notoriously difficult, and clustering learning only reported a 50\% accuracy overall with 2 layers.
All results are much lower than the current state of art in this dataset, which is close to 90\% \cite{ciresan_multi-column_2012}. But again we stress that our goal is real-time implementation and other researcher have obtained record results in this dataset only with very large networks, most of which are not amenable to real-time operation.

We tested random filters used on the first 2 layers of the networks for both SVHN and CIFAR10 datasets. The 1st layer of the network was trained with CL, as mentioned above. The randomly connected network used a fixed CNN layer as described in section \ref{sec-net-arch}.  Same qualitative results were obtained with a random CL model. The SpatialConvolution module was initialized with uniformly random filters by the Torch7 tool, and left untouched. 
We have found that random filters on the 2nd layer of this networks perform with the same levels of precision as the 2nd layers trained with CL. This was true for both SVNH and CIFAR10 datasets.

We also used the filters obtained with CL on a 2-layers standard CNN network. But the use of convolution operation was giving results that were 5-10\% less precise that the results obtained with sum-abs-diff operators.

\subsection{Dynamic datasets}

The results for the comparison for the real-time tracking network on the TLD dataset \cite{kalal_p-n_2010} are given in Table \ref{table1}.
Each video contains only one target. The metric used is the number of correctly tracked frames. More information and images on this dataset can be found here \cite{kalal_dataset_2010}. All network we tested were real-time networks performing at or above 5 frames/s.
As can be seen in Table \ref{table1}, the Clustering Learning (CL) network performs better than the convolutional neural network from \cite{farabet_scene_2012} in all sequences, and is comparable only in one sequence ("Pedestrian 2"). These results are currently not state-of-the-art, which is currently obtained in references \cite{dundar_visual_2012, kalal_tracking_2012}.

Not only the perfomance of the CL network is higher than CNN on the TLD dataset, but also the CNN from \cite{farabet_scene_2012} was trained in a week time, while the CL network was tried in 10 minutes. There is a clear advantage to using CL networks for general-purpose, dataset-free robotic vision tasks.

\begin{table}[htdp]
\caption{Precision comparison between of a Clustering Learning (CL) network  and a convolutional neural network (CNN)\cite{farabet_scene_2012} used as trackers in the TLD dataset \cite{kalal_tracking_2012}.}
\begin{center}
\begin{tabular}{|c|c|c|c|}
\hline\hline
Sequence 	& Frames 	& Precision: \textbf{CL}	& Precision: CNN \cite{farabet_scene_2012} \\ 
\hline
David 		& 761  	& \textbf{0.18} 	& 0.08  \\ 
Jumping 		& 313 	& \textbf{0.37} 	& 0.20 \\ 
Pedestrian 1 	& 140 	& \textbf{0.81} 	& 0.69 \\ 
Pedestrian 2 	& 338 	& 0.78 		& \textbf{0.79} \\ 
Pedestrian 3 	& 184 	& \textbf{0.45}  	& 0.44 \\ 
Car 			& 945 	& \textbf{0.67}  	& 0.48 \\ 
Carchase 		& 9928 	& \textbf{0.38} 	& 0.26 \\ 
\hline\hline
\end{tabular}
\end{center}
\label{table1}
\end{table}

In addition, we present in table \ref{table2} a comparison of running time of the most recent deep network work performing at the state-of-the-art in a variety of static datasets. We provide this comparison to show that although these networks perform extremely well on a specific dataset, we point out that similarly to the experimental results in \ref{table1}, that performance might not carry over to robotic vision tasks such as tracking of previously unseen objects.
We also want to point out that the state-of-the-art networks in table \ref{table2} are far from real-time operation, and it is not clear in their publication how that can be achieved down-sizing the networks or what performance they might attain.

\begin{table}[htdp]
\caption{Comparison of execution time of state-of-the-art networks as compared to the proposed CL network.}
\begin{center}
\begin{tabular}{|c|c|c|c|c|}
\hline\hline
Publication	&Frames/s 	&Precision	&1st layer filters	&Whitening\\
\hline
CL (this work)		&5-10	&89\%, SVHN		&16 			&none \\
\cite{farabet_scene_2012}	&1-2		&79.5\%, Stanford	&16 			&none \\
\cite{socher_convolutional_2012}	&1? 		&80.8\%, RGB-D	&128		&ZCA \\
\cite{coates_selecting_2011}	&0.1?		&82\%, CIFAR10	&1600		&ZCA \\
\cite{krizhevsky_imagenet_2012} &1?		&98.5\% ImageNet	&96			&none\\
\cite{ciresan_multi-column_2012}	&0.3?		&88.8\%, CIFAR10	&300		&none\\
\cite{jarrett_what_2009}	&2-4?		&99.5\%, MNIST	&32			&none\\
\hline\hline
\end{tabular}
\end{center}
\label{table2}
\end{table}

The data provided in table  \ref{table2} reports published and informally obtained data on the best results obtained in static datasets.  Cl networks also reported 50\% precision on CIFAR10, a low value compared to \cite{coates_selecting_2011, ciresan_multi-column_2012}. As reference \cite{socher_convolutional_2012} precision we used the RGB only data. Most published results did not report computer time. We asked the author to provide us the data for this table, when available. We estimated other network compute time based on the size of the 1st layer of their network, which is usually the most computationally demanding. We estimated frames/s based on the speed of our networks.  We want to stress the importance of reporting computing time for deep networks, and the use of whitening, as it is the only way to compare large and small networks and try to find the optimal size for robotic vision system, or other tasks.



\section{Discussion}

We presented results on clustering learning algorithms for general-purpose vision system. These algorithms can be used to train multi-layer feed-forward neural networks in minutes, with fully automated scripts with a reduced set of learning parameters. We show results on static datasets and dynamic tracking of objects in videos. We show results that prove that CL techniques are a viable and quick option to training deep networks. In static datasets, although they do not perform at the state-of-the-art because of their small network size, they can however be run in real-time because of the small network size. The accuracy gap in recognition can be reduced by using more features, as reported in \cite{coates_analysis_2011}, with an evaluation time linearly proportionalto the number of filters used. We also show that on tracking datasets the CL networks outperforms CNN networks. We believe the tracking dataset is a better approximation of robotic tasks, where locking on a target object is required for approach and manipulation.

We show that sum-abs-diff operators can be more effective than convolution operations for filtering in deep networks. We also point out that custom hardware with these operators will be more efficient in power and space.

Many groups in the field of deep learning work on static datasets to demonstrate the best learning techniques. A lot of these techniques cannot be applied in robotic vision system with the current modern hardware because they cannot perform in real-time ad require too much computational load.

We argue that more research is needed in the field of applied and real-time deep networks, where shortcuts need to be taken in order to optimize performance for speedy operation. The use of custom hardware is recommended \cite{farabet_neuflow_2011} but not necessary in many applications. 

CL models are also very interesting to bio-inspired vision research because they provide a close connection between computational neuroscience and machine learning. In particular unsupervised clustering algorithms provide a simplistic model of Hebbian Learning methods, where neurons that respond to the same input are clustered \cite{sanger_optimal_1989, foldiak_forming_1990, oja_principal_1992, bell_independent_1997} . Most of previous work was mathematically complex and not efficient to implement. In this paper we thus present one of the first practical application of Hebbian-like learning applied to deep networks. Its fast and simple computation can help researchers quickly train complex neural networks.

Further research is also needed to justify unsupervised learning on the 2nd or later layers. Our experiments indicate that random filters are as good as learned CL filters. This fact was also shown in other papers \cite{coates_selecting_2011,socher_convolutional_2012} where random filters were used in the 2nd layer, after one CL-trained layer with a large number of features (128 or above).

All the code for this paper is available here: \url{https://github.com/culurciello/CL_paper1_code}.

\subsubsection*{Acknowledgments}
We are especially grateful to the the Torch7 developing team, in particular Ronan Collobert, who first developed this great, easy and efficient tool, Clement Farabet, Koray Kavukcuoglu, Leon Bottou. We could not have done any of this work without standing on these giants shoulders.

\bibliography{clustering}
\bibliographystyle{unsrt}

\end{document}